\def\arxivversion{1}

\documentclass[letterpaper]{article} 

\ifdefined\arxivversion
  \usepackage{aaai2027}
\else
  \usepackage[submission]{aaai2027}  
\fi
\usepackage[hyphens]{url}  
\usepackage{graphicx} 
\usepackage{natbib}  
\usepackage{caption} 
\usepackage{microtype}
\usepackage{booktabs}
\usepackage{multirow}
\usepackage{amsmath}
\usepackage{amssymb}

\usepackage{xcolor}
\usepackage{colortbl}
\usepackage{enumitem}
\usepackage{algorithm}
\usepackage{algpseudocode}

\definecolor{gaterow}{RGB}{245, 238, 226}   
\definecolor{tagink}{RGB}{124, 79, 18}       

\title{FinReportBench: Measuring and Improving \\
       Institution-Grade Financial Report Generation}

\ifdefined\arxivversion
  \author{
    Yinghao Tang,\textsuperscript{\rm 1}
    Tan Zhenwei,\textsuperscript{\rm 2}
    Yiyao Wang,\textsuperscript{\rm 1}
    Wanli Gu,\textsuperscript{\rm 2}
    Xiaolu Zhang,\textsuperscript{\rm 2}
    JUN ZHOU,\textsuperscript{\rm 2}
    Wei Chen\textsuperscript{\rm 1}
  }
  \affiliations{
    \textsuperscript{\rm 1}State Key Laboratory of CAD\&CG, Zhejiang University \\
    \textsuperscript{\rm 2}Ant Group
  }
\else
  \author{
      Anonymous Authors
  }
  \affiliations{
      Anonymous Institution \\
      anonymous@example.com
  }
\fi

\begin{document}
\maketitle

\begin{abstract}
Large language models can produce fluent financial analysis, but
fluency does not establish whether a report is suitable for
institutional delivery. Existing evaluations use broad dimensions that
do not identify the specific defects that cause professional rejection.
We introduce \textbf{FinReportBench}, an expert-grounded benchmark for
measuring and improving institution-grade financial report generation.
In a pilot review, three experts assign all 75 reports from three models
the lowest readiness score despite generally fluent financial prose;
the recurring gaps concern report identity, institutional components,
source discipline, and visual delivery. Pilot interviews further
identify a hierarchy of report criteria. We
then derive the rubric through two rounds: expert partial orders over a
small output set guide item induction with multimodal evidence, and
experts audit the resulting decision boundaries. The final instrument
contains 35 observable items across deliverability, report identity,
and institutional completeness. Starting from 10,000 balanced Chinese
and English financial-research source records, we curate 244 bilingual
tasks across three research objects and two input tiers.
Each task separates the public query, reconstructed
research trajectory, and hidden source packet. Three independent judge
families reproduce the expert partial order at near-ceiling rates,
showing that the rubric's bounded, observable criteria support reliable
evaluation. Across nine model families, basic deliverability is nearly
saturated, while report identity and institutional completeness remain
the primary bottlenecks. The largest cross-model gaps arise in
generation-trace control, information density, and data discipline
rather than basic report framing. Finally, we use benchmark-guided skill
distillation to convert these recurrent failures into reusable
generation and self-review constraints. Across five model families, the
evolved skill improves mean G1 by 33.85 points and mean G2 by 13.83
points over the paired no-skill condition, while preserving G0 for
every pair.
\ifdefined\arxivversion
Code and benchmark artifacts are available at
\url{https://github.com/MisterBrookT/finreportbench}.
\fi
\end{abstract}

\section{Introduction}
\label{sec:intro}

Financial institutions communicate research through professional
reports. These reports combine analytical claims, data, charts, source
attribution, institutional identity, and a stable document structure
\citep{jin-etal-2026-finsight,lifan-etal-2026-cogito}.
They are reviewed, distributed, and archived as accountable research
artifacts. We refer to reports that satisfy this delivery standard as
\emph{institution-grade financial reports}.

Language models can produce fluent financial analysis, but their
outputs often fail to convince professional readers. In preliminary
interviews with three financial experts, all three assigned a
satisfaction score of 1, the lowest value on the five-point scale, to
all 75 reviewed model-generated artifacts.
They found that an output could contain plausible or locally accurate
statements while still falling far short of an acceptable research
deliverable. The largest deficits concerned professional report
conventions and visual presentation, including front-page framing,
institutional identity, page systems, compliance components, and
source presentation. These shortcomings call for a systematic
evaluation that can identify specific problems in generated financial
reports.

Recent work has begun to benchmark financial report generation, but
its evaluation criteria remain broad and coarse-grained. FinSight
evaluates factual accuracy, information effectiveness, and presentation
quality, while Cogito evaluates data quality, analytical quality, and
presentation quality
\citep{jin-etal-2026-finsight,lifan-etal-2026-cogito}. These dimensions
support overall system comparison, but they do not identify specific
problems that prevent institutional use. For example, a webpage-like
report with polished charts may receive a high presentation score even
when it lacks a publisher, analyst identity, and a stable page system.

To fill this gap, we present \textbf{FinReportBench}, the first
benchmark designed to measure whether generated financial reports meet
institutional delivery standards. Building such a benchmark requires
addressing three challenges. First, converting expert knowledge into a
reliable rubric is difficult. Professionals rely on conventions learned
through practice and can often express relative preferences more
reliably than an exhaustive scoring rule. Interviews and preference
labels must therefore be turned into observable criteria, clear
boundaries, and rules that another annotator can reproduce. Full-report
review is also costly, which limits the amount of expert annotation
available for rubric construction. Second, a published report shows
only the final artifact. The original client request and the research
steps that produced it are usually unavailable, so realistic benchmark
inputs cannot be collected directly. Third, a broad score can rank
systems but does not tell a developer what to fix. Actionable
evaluation requires specific failure evidence that can guide
improvement without exposing test answers or benchmark content.

\begin{figure*}[t]
  \centering
  \includegraphics[width=\textwidth]{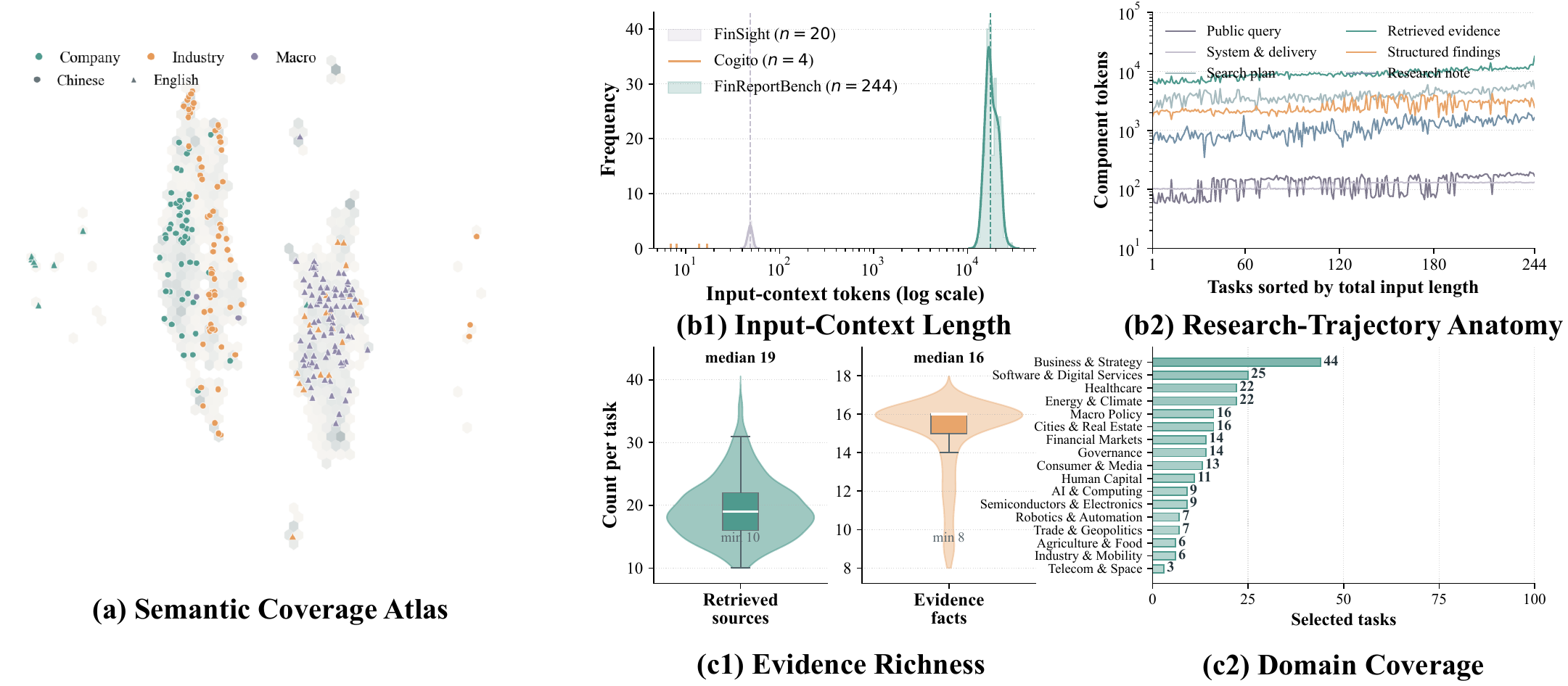}
  \caption{Metadata overview of FinReportBench. (a) A two-dimensional
  projection of the shared multilingual embedding space places the
  244 selected tasks across the density of the 10,000-task reference
  pool; color denotes research object and marker shape is a language
  diagnostic. Coverage is computed in the original embedding space,
  where the suite covers 95.65\% of reference tasks at cosine
  similarity $\geq 0.60$. (b1) Model inputs are substantially longer
  than query-only financial-report benchmarks; (b2) component profiles
  show the anatomy of the 244 scripted research trajectories. (c1)
  Each task contains multiple retrieved sources and structured evidence
  facts. (c2) Counts across the diagnostic domain taxonomy show the
  breadth of selected research topics.}
  \label{fig:dataset-metadata}
\end{figure*}

We make five contributions:
\begin{itemize}[leftmargin=*,itemsep=3pt]
  \item \textbf{Expert-grounded gap diagnosis.} We provide an
  expert-grounded diagnosis of why fluent model outputs remain
  unsuitable as institutional reports. Reviews with three senior
  financial professionals identify the gap and reveal a hierarchy of
  professional requirements (\S~2).
  \item \textbf{Financial-report generation dataset.} We introduce a
  244-task bilingual dataset curated from a balanced 10,000-task reference
  space, covering three research objects, two input tiers, and 95.65\%
  of the multilingual semantic space. Each task includes a tool-use
  research trajectory with separated public queries, retrieved
  evidence, and hidden source packets (\S~3).
  \item \textbf{Hierarchical evaluation framework.} We propose a
  hierarchical framework derived from expert reviews and
  preference-guided rubric induction. It evaluates institutional
  readiness through 35 observable criteria and an evidence-based
  multimodal judging protocol (\S~4).
  \item \textbf{Benchmark-guided skill evolution.} We propose a method
  that converts recurrent item-level failures across
  models into a compact, reusable production skill. The method develops
  and selects skill revisions on external tasks before a locked
  cross-model transfer test (\S~5).
  \item \textbf{Evaluation and findings.} In the completed evaluation,
  we characterize nine model families through item-level analyses on
  FinReportBench. The same task set is used for every model, and visible
  output defects remain part of the measured result.
  We also test whether the evolved skill transfers across
  five model families (\S~6).
\end{itemize}

\section{Related Work and Pilot Interviews}
\label{sec:background-interviews}

\subsection{Related Work}

Financial NLP benchmarks largely evaluate question answering and
multimodal reasoning over tables, filings, and conversations
\citep{chen2021finqa,chen2022convfinqa,zhu2021tatqa,
zhao2022multihiertt,islam2023financebench,reddy2024docfinqa,
liu2025finmme,xie2024finben,xie2023pixiu,wu2023bloomberggpt}, whereas
agent benchmarks and training methods emphasize web interaction,
retrieval, and multi-step information seeking
\citep{mialon2023gaia,zhou2023webarena,wu2025webdancer,
nakano2021webgpt,jin2025searchr1,wei2025browsecomp,
deng2023mind2web,gou2025mind2web2,baek2024researchagent,
li2026deepeye,bian2025you}. Recent
financial-report systems add
broad factual, analytical, and presentation dimensions
\citep{jin-etal-2026-finsight,lifan-etal-2026-cogito,
weng2025datalab,tang2026vividoc,xie2024haichart,xie2026datamagic,
tan2024proxyqa,
wu2024longgenbench,kim2024biggen,asai2024openscholar}, but these scores
do not identify the specific defects that cause professional rejection.
FinReportBench instead evaluates complete rendered reports through
fine-grained observable items organized by an expert-derived hierarchy.

Recent work also extracts reusable agent skills from resources or
execution failures and selects revisions on held-out tasks
\citep{li2026skillsbench,huang2026skilllens,alzubi2026evoskill,
shen2026skillfoundry,zhang2026coevoskills}. We adapt this idea to
institutional reporting: recurrent item-level failures become compact
production rules, and external validation guards against negative
transfer.

\subsection{Pilot Expert Interviews}

We conducted multi-round pilot interviews with three senior financial
professionals, each with more than ten years of experience. They
reviewed 75 anonymized reports produced by three models for 25 queries
and rated complete-report readiness on a five-point scale.
\textbf{All 75 reports received the lowest score of 1.} The interviews
yielded three design findings. First, fluent outputs still lacked
professional structure, institutional identity, source discipline,
risk disclosure, and consistent visual delivery
\citep{qu2024finflier,wen2026multimodal,tang2026chartplot}. Second,
review was
hierarchical: experts checked deliverability, then report identity,
then detailed institutional components. Third, research preceded
writing; analysts expected a model to continue from an organized
evidence state rather than generate directly from a short query.

These findings motivate, respectively, fine-grained item evaluation,
the G0--G1--G2 hierarchy, and the reconstructed research trajectory
used by FinReportBench. Interview questions, review instructions, and
additional protocol details appear in the supplementary material.

\section{FinReportBench Dataset}
\label{sec:benchmark}

\subsection{Dataset Design Objectives}

FinReportBench targets four properties: \textbf{authenticity}, by
deriving tasks from real analyst or institutional sources;
\textbf{diversity}, across languages, research objects, domains, and
query specificity; \textbf{efficiency}, by selecting a compact suite
that preserves the semantic coverage of a much larger pool; and a
\textbf{realistic generation pipeline}, in which report writing
continues from an evidence-rich research trajectory rather than a
query alone.

\subsection{Construction Pipeline}

\begin{figure*}[t]
  \centering
  \includegraphics[width=0.85\textwidth]{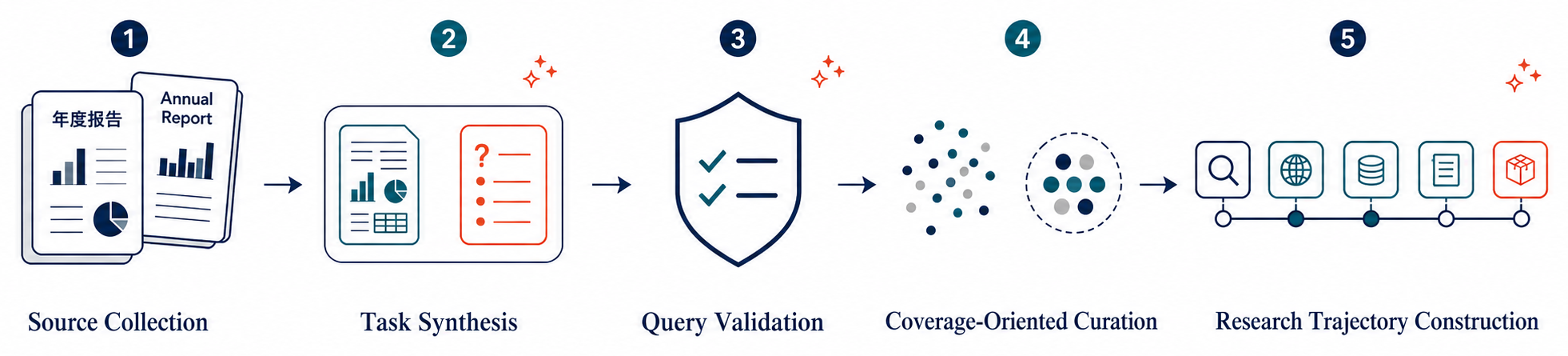}
  \caption{The FinReportBench dataset construction pipeline. We
  synthesize and validate source-grounded tasks, curate the benchmark
  for broad coverage, and construct a research trajectory for each
  selected task.}
  \label{fig:dataset-pipeline}
\end{figure*}

Figure~\ref{fig:dataset-pipeline} summarizes five stages: source
collection, task synthesis, query validation, coverage-oriented
curation, and trajectory construction.

\paragraph{Source-grounded task synthesis.}
We assemble a balanced multilingual reference pool of 5,000 Chinese
analyst reports from Eastmoney and 5,000 English institutional
publications from the World Bank. Because original client requests are
rarely available, reverse synthesis recovers a plausible request from
each source's research object, context, and analytical angle. An
LLM-based audit rejects unnatural requests, taxonomy errors, and
publisher, source, or answer leakage. Each accepted task separates a
public query card from a hidden source mapping and construction
metadata; the source artifact is never provided to the report model.

\paragraph{Coverage-oriented curation.}
We embed all source-derived research tasks in one multilingual semantic
space and select a compact suite while balancing language, research
object, domain, and query specificity. Language is a diagnostic slice,
not a separate benchmark. The resulting 244 tasks contain 122 Chinese
and 122 English requests; 61 are company, 92 industry, and 91 macro
tasks. Eighty-seven are open-ended T0 requests and 157 are
thesis-guided T1 requests. At cosine similarity 0.60, the selected set
covers 95.65\% of the 10,000-task reference space, including 99.46\%
of Chinese and 91.84\% of English references.

\paragraph{Research trajectory construction.}
\label{sec:trajectory}
For every selected task, we reconstruct a scripted deep-research
trajectory containing the public request, search decisions, retrieved
evidence, and a final textual research note. The final user message
asks for a self-contained HTML report. The trajectory does not replay
the hidden source and excludes benchmark identifiers, internal evidence
labels, and URL lists from the research note. All 244 trajectories pass
role-order, provenance, final-note, and leakage checks.

The supplementary material provides source-pool composition, semantic
fingerprint fields, embedding and selection details, query-audit
criteria, the complete trajectory contract, and additional dataset
statistics.

\section{Evaluation Design}
\label{sec:eval}

\subsection{Evaluation Design Objectives}

Our evaluator targets three properties. \textbf{Low-cost expert
grounding} uses sparse within-query preferences instead of exhaustive
item annotation. \textbf{Fine-grained diagnosis} returns observable
item decisions and supporting evidence rather than only an overall
score. \textbf{Hierarchical validity} follows the professional review
order found in the pilot study, preventing later strengths from fully
compensating for an earlier-stage failure.

\subsection{Evaluation Construction Pipeline}

\begin{figure*}[t]
  \centering
  \includegraphics[width=0.9\textwidth]{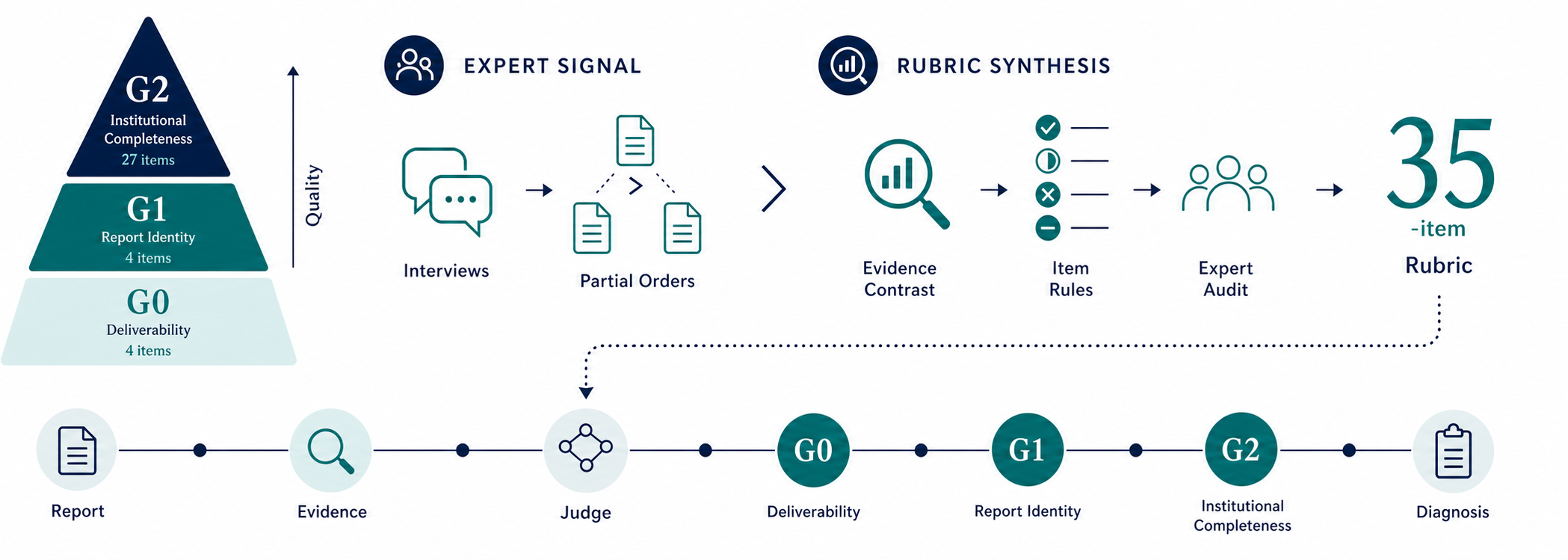}
  \caption{The FinReportBench evaluation pipeline. Sparse expert
  preferences guide contrastive evidence mining and observable item
  induction. Expert-audited items form a hierarchical rubric that
  returns layer scores, item decisions, and supporting evidence from
  rendered pages and extracted text.}
  \label{fig:evaluation-design}
\end{figure*}

Figure~\ref{fig:evaluation-design} summarizes the process. Experts first
provide partial orders over small sets of anonymized reports generated
for the same query. A multimodal model contrasts preferred and
dispreferred reports to mine candidate visual and textual evidence
\citep{tang2026igenbench,pan2025visshepherd,masry2022chartqa,
kantharaj2022opencqa,methani2020plotqa,masry2023unichart,
xie2025visjudge,chen2025chartmark}.
Recurring, generalizable differences are converted into observable
items with \textit{pass}, \textit{partial}, \textit{fail}, and, where
appropriate, \textit{not-applicable} boundaries. Experts then remove,
merge, split, or revise invalid and redundant items. Rubric induction
and final validation use disjoint report samples.

This two-round process concentrates scarce expert effort on
decision-relevant contrasts: experts supply preference boundaries and
audit induced items rather than label every report--criterion pair.
The supplementary material contains the annotation protocol,
contrastive mining procedure, item-induction prompts, and additional
audit examples.

\subsection{Hierarchical Rubric and Scoring}

The frozen rubric contains 35 items in three ordered layers.
\textbf{G0: deliverability} uses four preflight checks for a visible,
readable, continuous, and non-broken artifact; any required failure
sets the total to zero. \textbf{G1: report identity} uses four items to
test whether the artifact is recognizable as institutional research
rather than a webpage, dashboard, slide deck, generic summary, or
direct task answer. \textbf{G2: institutional completeness} uses 27
items covering front-page framing, institutional identity, compliance,
page systems, information density, source and chart discipline, and
generation-artifact control.

For active G1 and G2 items, pass receives 1, partial 0.5, and fail 0;
not applicable is removed from the denominator only when a criterion
genuinely does not apply. Let $S_1$ be the weighted G1 percentage, with
$g_1,d_1$ the weighted points obtained and available in G1, and define
$g_2,d_2$ analogously for G2. For a G0-passing report,
\begin{equation}
  S = 100\frac{g_1 + (S_1/100)g_2}{d_1+d_2};
  \qquad S=0\ \text{if G0 fails}.
  \label{eq:hierarchical-score}
\end{equation}
Thus weak report identity discounts isolated G2 details without a
discontinuous score cap. We report G1 and G2 alongside this provisional
total. Complete item descriptors, weights, boundaries, and scoring
examples appear in the supplementary material.

\subsection{Itemized Multimodal Evaluation}

We render each HTML report and extract its text. The judge evaluates
every applicable item using the appropriate visual, textual, or mixed
evidence and returns an item decision with a short evidence citation.
This produces both comparable layer scores and an actionable failure
profile for model or skill improvement.

Held-out validation compares automatic and expert within-query partial
orders over model outputs. We also compare judge families while fixing
reports, rubric, and scoring code
\citep{zheng2023mtbench,chiang2024arena,tan2024judgebench,
zhuge2024agentjudge,zheng2024judging}. Section~\ref{sec:experiments} reports
the main validation results; complete pair-sampling instructions,
prompts, and additional calibration matrices are supplied in the
supplementary material.

\section{Benchmark-Grounded Skill Evolution}
\label{sec:skill}

\subsection{Skill Design Objectives}

Our method follows three objectives. \textbf{Diagnostic grounding}
requires every revision to address recurring, observable item-level
failures rather than unconstrained self-reflection.
\textbf{Generalizable improvement} favors compact rules that transfer
across queries and model families instead of fitting one output.
\textbf{Leakage-safe evolution} proposes and selects revisions only on
external cases; the full FinReportBench test set is used after the skill
is frozen. Model weights remain unchanged throughout.

\subsection{Evolution Pipeline}

\begin{figure}[t]
  \centering
  \includegraphics[width=\columnwidth]{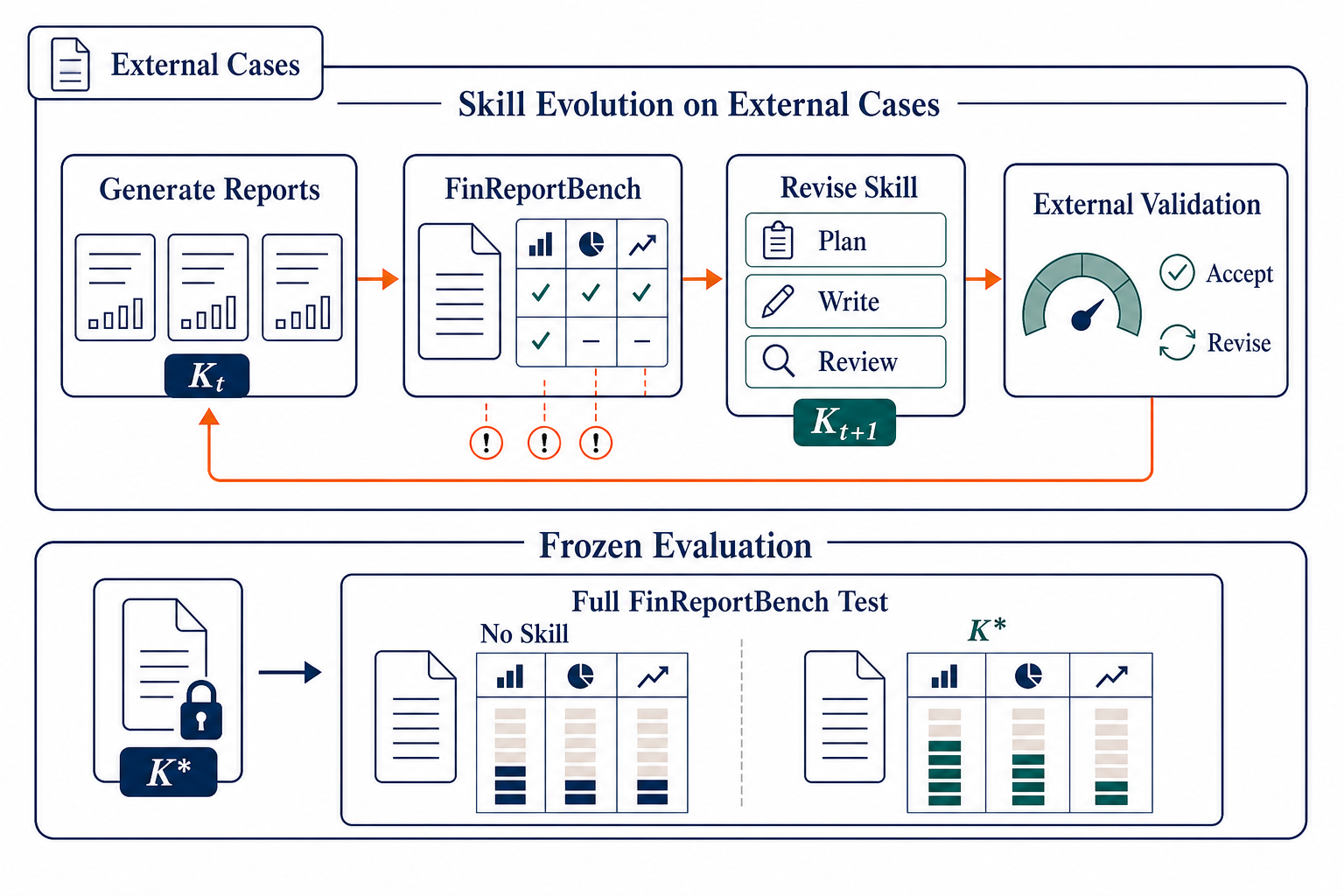}
  \caption{Benchmark-grounded skill evolution. External cases support
  iterative diagnosis, revision, and validation; the selected skill is
  frozen before paired evaluation on FinReportBench.}
  \label{fig:skill-evolution}
\end{figure}

Figure~\ref{fig:skill-evolution} shows the complete pipeline. Starting
from skill $K_t$, several report models generate outputs for external
financial-report cases. FinReportBench converts their rendered reports
into item-level decisions and observable evidence. An optimizer turns
recurring failures into a candidate $K_{t+1}$, organized as planning,
writing, and review rules. Query-disjoint external validation either
accepts the candidate or retains $K_t$. When improvement stops, the
best accepted skill is frozen as $K^\star$ and evaluated once on a
locked benchmark subset against a paired no-skill condition.

The evolution and validation sets cover the same broad financial-report
setting but share no query, trajectory, or source document with the
benchmark. We use five external discovery cases and five disjoint
external validation cases, balanced across Chinese and English and
covering company, industry, and macro research. Using multiple report
models exposes failures that recur across model families while holding
each model's generation settings fixed.

\subsection{Failure-Grounded Skill Revision}

For each external report, the evaluator returns a categorical decision
and short visual or textual evidence for every applicable item. We
aggregate outcomes across cases and models and revise only failures
with sufficient cross-model support. The optimizer receives item
semantics, aggregate outcomes, and anonymized evidence, but not item
identifiers, weights, judge prompts, benchmark queries, or hidden source
packets.

The optimizer contrasts failed and passed evidence and writes compact
constraints with three parts: when a rule applies, what the generator
should do, and how the output can be checked. The resulting skill has
\emph{plan}, \emph{write}, and \emph{review} blocks. A critic removes
task-specific facts, rubric wording, duplication, and conflicting
instructions, and rejects rules that require unsupported analyst
identities, qualifications, ratings, or target prices. This keeps the
skill actionable without copying the rubric into a long prompt.

\subsection{External Validation and Freezing}

Each candidate is tested on held-out external cases. It is accepted
only when targeted items improve without reducing deliverability,
report identity, institutional completeness, or materially regressing
non-target items; otherwise the previous skill remains active. This
failure-driven update follows the general pattern of
\citet{alzubi2026evoskill}, but uses rendered financial reports and
hierarchical item evidence as feedback.

Evolution stops after a fixed round budget or a patience window without
validation improvement. The best accepted revision, rather than the
last proposal, becomes $K^\star$. Only then does each report model
receive the locked FinReportBench evaluation under paired no-skill and
$K^\star$ conditions, with query, trajectory, generation budget, and
inference settings held fixed. Test queries and their order are frozen
before generation; selection does not use report or judge outcomes. We
measure changes in G0, G1, G2, item pass rates, the provisional total,
non-target regressions, generation cost, and skill length.

Exact external-case composition, support and acceptance thresholds,
stopping parameters, skill budget, optimizer and critic prompts, and
pseudocode are provided in the supplementary material.

\section{Experiments}
\label{sec:experiments}

\subsection{Setup}

The completed run evaluates nine recent model families from
international and Chinese providers: DeepSeek V4 Flash, DeepSeek V4
Pro, Qwen 3.7 Max, GLM-5.2, Kimi K2.6, MiniMax M2.5, and MiniMax M2.7.
The run additionally includes MiniMax M3 and Qwen 3.6 27B. Every system
is evaluated on FinReportBench in the \texttt{no\_skill} condition. We
record model identifiers, run date, and inference settings in the
supplementary material.
We render every output and evaluate extracted text together with up to
four rendered pages using GPT-5.6 Luna at medium reasoning effort and
the frozen 35-item rubric. All systems share the same judge prompt,
rubric, and scoring code. Tasks, trajectories, and the output contract
are also fixed before the run.

\subsection{Main Benchmark Results}

Every system is evaluated on the same benchmark tasks, while visible
generation and encoding defects remain system errors. The provisional
hierarchical-score ordering is MiniMax M2.7 (22.6),
GLM-5.2 (21.8), MiniMax M2.5 (21.1), DeepSeek V4 Flash (20.2),
Qwen 3.7 Max (19.1), DeepSeek V4 Pro (18.8), MiniMax M3 (18.6),
Kimi K2.6 (17.7), and Qwen 3.6 27B (15.1). MiniMax M2.7 has the
highest G1 score (40.0) and G2 score (46.6).
Because the total remains provisional, we report G1 and G2 alongside
it and do not interpret small score gaps as stable ability differences.
Table~\ref{tab:main-results} presents every rubric item and the
provisional total.

\definecolor{resultgzero}{RGB}{58,139,130}
\definecolor{resultgone}{RGB}{220,142,72}
\definecolor{resultgtwo}{RGB}{126,111,157}
\definecolor{resultoverall}{RGB}{105,119,126}
\def\resultcellcolor{resultoverall}
\newcommand{\itemcell}[1]{%
  \ifdim #1pt > 80pt\expandafter\cellcolor\expandafter{\resultcellcolor!32}#1%
  \else\ifdim #1pt > 60pt\expandafter\cellcolor\expandafter{\resultcellcolor!24}#1%
  \else\ifdim #1pt > 40pt\expandafter\cellcolor\expandafter{\resultcellcolor!16}#1%
  \else\ifdim #1pt > 20pt\expandafter\cellcolor\expandafter{\resultcellcolor!9}#1%
  \else\expandafter\cellcolor\expandafter{\resultcellcolor!4}#1%
  \fi\fi\fi\fi}

\begin{table*}[t]
  \centering
  \caption{Item-level FinReportBench results across all 35 rubric items.
  Every system is evaluated on the same benchmark tasks, and visible output defects remain included.
  Cells report mean item credit (pass=100, partial=50, fail=0).
  Overall is the provisional hierarchical score.}
  \label{tab:main-results}
  \tiny
  \setlength{\tabcolsep}{0.7pt}
  \renewcommand{\arraystretch}{1.10}
  \resizebox{\textwidth}{!}{%
  \begin{tabular}{@{}l<{\gdef\resultcellcolor{resultgzero}}
    rrrr<{\gdef\resultcellcolor{resultgone}}@{\hspace{3pt}}
    rrrr<{\gdef\resultcellcolor{resultgtwo}}@{\hspace{3pt}}
    *{26}{r}r<{\gdef\resultcellcolor{resultoverall}}@{\hspace{3pt}}
    r@{}}
    \toprule
    & \multicolumn{4}{c}{\textbf{G0: Deliverability}}
    & \multicolumn{4}{c}{\scalebox{0.72}{\textbf{G1: Report identity}}}
    & \multicolumn{27}{c}{\textbf{G2: Institutional completeness}}
    & \textbf{Overall} \\
    \cmidrule(lr){2-5}\cmidrule(lr){6-9}\cmidrule(lr){10-36}\cmidrule(l){37-37}
    \textbf{System} & \rotatebox{65}{\textbf{D01}} & \rotatebox{65}{\textbf{D02}} & \rotatebox{65}{\textbf{D03}} & \rotatebox{65}{\textbf{D04}} & \rotatebox{65}{\textbf{R01}} & \rotatebox{65}{\textbf{R02}} & \rotatebox{65}{\textbf{R03}} & \rotatebox{65}{\textbf{R04}} & \rotatebox{65}{\textbf{C01}} & \rotatebox{65}{\textbf{C02}} & \rotatebox{65}{\textbf{C03}} & \rotatebox{65}{\textbf{C04}} & \rotatebox{65}{\textbf{C05}} & \rotatebox{65}{\textbf{C06}} & \rotatebox{65}{\textbf{C07}} & \rotatebox{65}{\textbf{C08}} & \rotatebox{65}{\textbf{C09}} & \rotatebox{65}{\textbf{C10}} & \rotatebox{65}{\textbf{C11}} & \rotatebox{65}{\textbf{C12}} & \rotatebox{65}{\textbf{C13}} & \rotatebox{65}{\textbf{C14}} & \rotatebox{65}{\textbf{C15}} & \rotatebox{65}{\textbf{C16}} & \rotatebox{65}{\textbf{C17}} & \rotatebox{65}{\textbf{C18}} & \rotatebox{65}{\textbf{C19}} & \rotatebox{65}{\textbf{C20}} & \rotatebox{65}{\textbf{C21}} & \rotatebox{65}{\textbf{C22}} & \rotatebox{65}{\textbf{C23}} & \rotatebox{65}{\textbf{C24}} & \rotatebox{65}{\textbf{C25}} & \rotatebox{65}{\textbf{C26}} & \rotatebox{65}{\textbf{C27}} & \rotatebox{65}{\textbf{All}} \\
    \midrule
    MiniMax M2.7 & \itemcell{100} & \itemcell{99} & \itemcell{100} & \itemcell{100} & \itemcell{100} & \itemcell{1} & \itemcell{51} & \itemcell{22} & \itemcell{56} & \itemcell{100} & \itemcell{86} & \itemcell{51} & \itemcell{96} & \itemcell{7} & \itemcell{6} & \itemcell{50} & \itemcell{0} & \itemcell{0} & \itemcell{0} & \itemcell{47} & \itemcell{0} & \itemcell{66} & \itemcell{0} & \itemcell{3} & \itemcell{18} & \itemcell{100} & \itemcell{100} & \itemcell{90} & \itemcell{82} & \itemcell{100} & \itemcell{67} & \itemcell{50} & \itemcell{85} & \itemcell{31} & \itemcell{18} & \itemcell{22.6} \\
    GLM-5.2 & \itemcell{100} & \itemcell{99} & \itemcell{100} & \itemcell{100} & \itemcell{100} & \itemcell{0} & \itemcell{51} & \itemcell{21} & \itemcell{2} & \itemcell{100} & \itemcell{91} & \itemcell{51} & \itemcell{98} & \itemcell{5} & \itemcell{8} & \itemcell{50} & \itemcell{0} & \itemcell{0} & \itemcell{0} & \itemcell{51} & \itemcell{0} & \itemcell{77} & \itemcell{0} & \itemcell{4} & \itemcell{10} & \itemcell{100} & \itemcell{100} & \itemcell{95} & \itemcell{89} & \itemcell{100} & \itemcell{51} & \itemcell{50} & \itemcell{88} & \itemcell{23} & \itemcell{3} & \itemcell{21.8} \\
    MiniMax M2.5 & \itemcell{100} & \itemcell{100} & \itemcell{100} & \itemcell{100} & \itemcell{100} & \itemcell{0} & \itemcell{49} & \itemcell{19} & \itemcell{54} & \itemcell{99} & \itemcell{84} & \itemcell{51} & \itemcell{88} & \itemcell{2} & \itemcell{3} & \itemcell{50} & \itemcell{0} & \itemcell{0} & \itemcell{0} & \itemcell{47} & \itemcell{0} & \itemcell{67} & \itemcell{0} & \itemcell{0} & \itemcell{9} & \itemcell{100} & \itemcell{100} & \itemcell{85} & \itemcell{77} & \itemcell{100} & \itemcell{45} & \itemcell{50} & \itemcell{97} & \itemcell{46} & \itemcell{41} & \itemcell{21.1} \\
    DeepSeek V4 Flash & \itemcell{100} & \itemcell{100} & \itemcell{100} & \itemcell{100} & \itemcell{100} & \itemcell{0} & \itemcell{52} & \itemcell{10} & \itemcell{78} & \itemcell{99} & \itemcell{83} & \itemcell{50} & \itemcell{96} & \itemcell{7} & \itemcell{2} & \itemcell{50} & \itemcell{0} & \itemcell{0} & \itemcell{0} & \itemcell{50} & \itemcell{0} & \itemcell{86} & \itemcell{0} & \itemcell{0} & \itemcell{11} & \itemcell{100} & \itemcell{100} & \itemcell{84} & \itemcell{84} & \itemcell{100} & \itemcell{51} & \itemcell{50} & \itemcell{92} & \itemcell{37} & \itemcell{5} & \itemcell{20.2} \\
    Qwen 3.7 Max & \itemcell{100} & \itemcell{100} & \itemcell{99} & \itemcell{100} & \itemcell{99} & \itemcell{0} & \itemcell{50} & \itemcell{12} & \itemcell{19} & \itemcell{99} & \itemcell{87} & \itemcell{50} & \itemcell{95} & \itemcell{5} & \itemcell{3} & \itemcell{52} & \itemcell{0} & \itemcell{0} & \itemcell{0} & \itemcell{49} & \itemcell{0} & \itemcell{61} & \itemcell{0} & \itemcell{1} & \itemcell{13} & \itemcell{100} & \itemcell{100} & \itemcell{87} & \itemcell{83} & \itemcell{100} & \itemcell{48} & \itemcell{49} & \itemcell{75} & \itemcell{32} & \itemcell{1} & \itemcell{19.1} \\
    DeepSeek V4 Pro & \itemcell{100} & \itemcell{100} & \itemcell{100} & \itemcell{100} & \itemcell{100} & \itemcell{0} & \itemcell{48} & \itemcell{9} & \itemcell{65} & \itemcell{99} & \itemcell{82} & \itemcell{50} & \itemcell{96} & \itemcell{1} & \itemcell{1} & \itemcell{50} & \itemcell{0} & \itemcell{0} & \itemcell{0} & \itemcell{49} & \itemcell{0} & \itemcell{83} & \itemcell{0} & \itemcell{0} & \itemcell{3} & \itemcell{100} & \itemcell{100} & \itemcell{83} & \itemcell{88} & \itemcell{99} & \itemcell{65} & \itemcell{50} & \itemcell{61} & \itemcell{27} & \itemcell{6} & \itemcell{18.8} \\
    MiniMax M3 & \itemcell{100} & \itemcell{100} & \itemcell{97} & \itemcell{100} & \itemcell{100} & \itemcell{1} & \itemcell{50} & \itemcell{11} & \itemcell{2} & \itemcell{100} & \itemcell{88} & \itemcell{50} & \itemcell{96} & \itemcell{2} & \itemcell{3} & \itemcell{50} & \itemcell{0} & \itemcell{0} & \itemcell{0} & \itemcell{51} & \itemcell{0} & \itemcell{52} & \itemcell{0} & \itemcell{0} & \itemcell{9} & \itemcell{100} & \itemcell{100} & \itemcell{92} & \itemcell{88} & \itemcell{100} & \itemcell{52} & \itemcell{50} & \itemcell{79} & \itemcell{42} & \itemcell{15} & \itemcell{18.6} \\
    Kimi K2.6 & \itemcell{100} & \itemcell{100} & \itemcell{100} & \itemcell{100} & \itemcell{100} & \itemcell{0} & \itemcell{50} & \itemcell{4} & \itemcell{32} & \itemcell{100} & \itemcell{89} & \itemcell{52} & \itemcell{91} & \itemcell{1} & \itemcell{1} & \itemcell{50} & \itemcell{0} & \itemcell{0} & \itemcell{0} & \itemcell{48} & \itemcell{0} & \itemcell{60} & \itemcell{0} & \itemcell{0} & \itemcell{3} & \itemcell{100} & \itemcell{100} & \itemcell{75} & \itemcell{89} & \itemcell{100} & \itemcell{46} & \itemcell{50} & \itemcell{95} & \itemcell{39} & \itemcell{25} & \itemcell{17.7} \\
    Qwen 3.6 27B & \itemcell{94} & \itemcell{95} & \itemcell{82} & \itemcell{100} & \itemcell{97} & \itemcell{0} & \itemcell{47} & \itemcell{10} & \itemcell{48} & \itemcell{98} & \itemcell{79} & \itemcell{47} & \itemcell{87} & \itemcell{4} & \itemcell{0} & \itemcell{50} & \itemcell{0} & \itemcell{0} & \itemcell{0} & \itemcell{43} & \itemcell{0} & \itemcell{47} & \itemcell{0} & \itemcell{0} & \itemcell{3} & \itemcell{97} & \itemcell{91} & \itemcell{82} & \itemcell{86} & \itemcell{95} & \itemcell{52} & \itemcell{50} & \itemcell{80} & \itemcell{47} & \itemcell{39} & \itemcell{15.1} \\
    \bottomrule
  \end{tabular}%
  }
\end{table*}

\subsection{Human--Evaluator Alignment}

We conduct a calibration-scale expert validation using relative
judgments rather than exhaustive item-level gold labels. Three senior
financial experts jointly review 18 anonymized reports from three
queries and establish a consensus three-tier partial order: three
authentic source reports rank above two generated reports judged to
have recognizable report form, which in turn rank above the remaining
13 generations. Expanding only cross-tier relations yields 71 directed
pairwise constraints. For each evaluator run, a constraint is satisfied
when the hierarchical score of the expert-preferred report is strictly
higher.

\begin{figure}[htbp]
  \centering
  \includegraphics[width=\columnwidth]{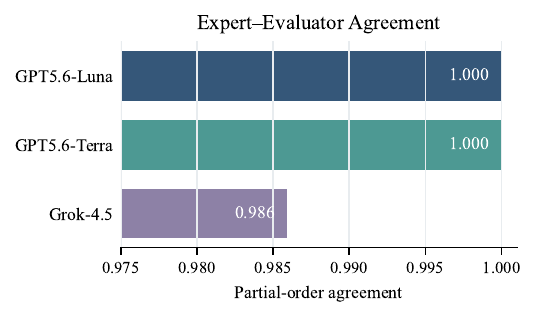}
  \caption{Alignment with the three-expert consensus partial order.
  GPT5.6-Luna and GPT5.6-Terra satisfy all 71 constraints, while
  Grok-4.5 satisfies 70. The axis is truncated to distinguish the
  near-ceiling results.}
  \label{fig:human-alignment}
\end{figure}

All three evaluator families preserve 98.6--100.0\% of the expert
partial-order constraints. A separate Terra repeat agrees with the
reference run on 97.0\% of item decisions, while Luna and Grok agree
with the Terra runs on 92.1--93.8\%; corresponding score-ranking
correlations range from 0.949 to 0.981. Together, the expert alignment
and cross-family agreement support the criterion validity and
reliability of the automatic evaluator.

\subsection{Hierarchical Diagnostic Profiles}

\begin{figure}[t]
  \centering
  \includegraphics[width=0.72\columnwidth]{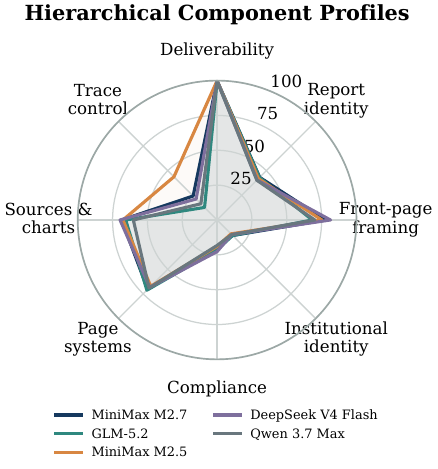}
  \caption{Hierarchical component profiles for five representative
  models. The radial axes aggregate related rubric items while
  preserving the G0--G1--G2 interpretation.}
  \label{fig:hierarchical-profiles}
\end{figure}

The leaderboard establishes ranking; the hierarchical profile explains
how systems differ. We aggregate the 35 items into deliverability,
report identity, front-page framing, institutional identity,
compliance, page systems, source and chart discipline, and
generation-trace control. Figure~\ref{fig:hierarchical-profiles} shows
five representative models in the main paper; profiles for all models
will appear in the supplementary material.

Cross-model gaps are largest for generation-trace control (31.1
points), followed by information density (13.4) and data discipline
(13.2). The report-frame gap is only 4.6 points.

\subsection{Performance Across Benchmark Slices}

\begin{figure}[t]
  \centering
  \includegraphics[width=\columnwidth]{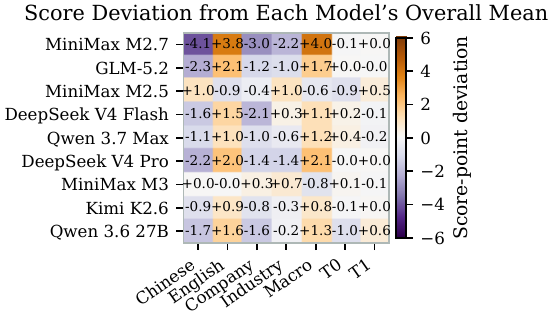}
  \caption{Performance variation across language, research object, and
  query-specificity slices. Cells show deviation from each model's
  overall score.}
  \label{fig:slice-analysis}
\end{figure}

We next test whether conclusions depend on one language or research
setting. We report Chinese versus English, company versus industry
versus macro, and open-ended T0 versus thesis-guided T1 requests.
Figure~\ref{fig:slice-analysis} displays each slice as a deviation from
the same model's overall score, preventing globally strong systems from
dominating the visual scale. Absolute scores, sample sizes, and
confidence intervals remain available in the supplementary tables.

We use these slices diagnostically rather than define separate
benchmarks. The largest model-relative slice deviation is 4.1 points.
Rankings are highly stable across T0 and T1 requests
($\rho=0.983$ and $1.000$) and remain similar across company,
industry, and macro tasks ($\rho=0.833$, $0.833$, and $0.900$).
Language is the largest source of rank variation
($\rho=0.650$ for Chinese and $0.917$ for English).

\subsection{Item Difficulty and Discrimination}

\begin{figure}[t]
  \centering
  \includegraphics[width=\columnwidth]{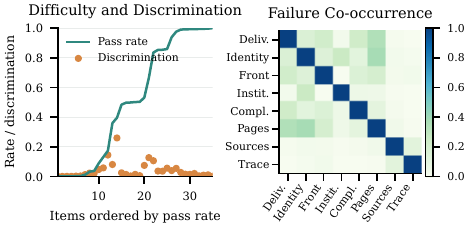}
  \caption{Item analysis. Left: pass rate and model discrimination for
  all 35 items. Right: failure co-occurrence aggregated by rubric
  component.}
  \label{fig:item-analysis}
\end{figure}

A useful benchmark should contain neither only trivial checks nor only
unattainable requirements. We therefore measure each item's pass rate
and model-discrimination strength, then cluster item failures by their
co-occurrence across reports. Figure~\ref{fig:item-analysis} separates
item difficulty from discrimination and reveals whether recurring
failures form coherent institutional components.

Mean item credit spans 0.0--99.9\%, confirming that the rubric contains
both near-universal delivery checks and consistently absent
institutional requirements. The maximum between-model standard
deviation is 25.9 points. Table~\ref{tab:main-results} provides the
complete auditable item matrix.

\subsection{Skill-Evolution Evaluation}

Finally, we test whether benchmark diagnostics can improve generation.
Five external discovery cases produce the initial skill $K_0$ and
candidate revisions; five disjoint external validation cases determine
which candidates are retained. After $K^\star$ is frozen, a locked
FinReportBench evaluation compares no skill and $K^\star$ across Qwen
3.7 Max, DeepSeek V4 Pro, DeepSeek V4 Flash, MiniMax M3, and GLM-5.2
under 100 paired task--model comparisons.

\begin{figure}[t]
  \centering
  \includegraphics[width=\columnwidth]{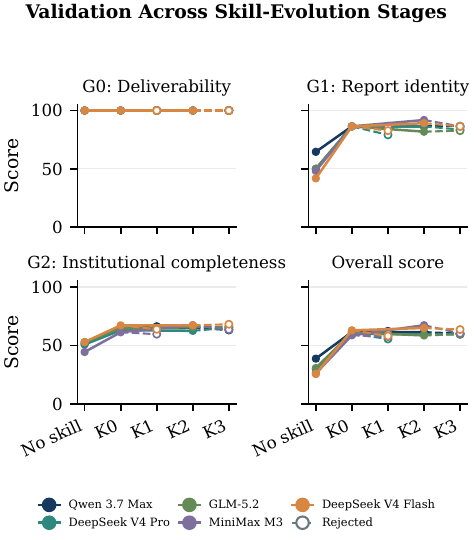}
  \caption{Validation performance across skill-evolution stages for
  five report models.
  Filled markers are retained versions and open markers are rejected
  candidates.}
  \label{fig:skill-evolution-results}
\end{figure}

On the locked FinReportBench evaluation, $K^\star$ improves mean G1 by
33.85 points (95\% CI: 31.17--36.63) and mean G2 by 13.83 points
(12.65--15.02), while preserving G0 for every pair. All five models
improve: G1 gains range from 26.39 to 41.86 points and G2 gains from
11.22 to 16.98 points. The larger G1 gain shows that the skill primarily
repairs report identity and document structure, while the consistent G2
gain shows that it also strengthens institutional completeness. The
shared direction across model families demonstrates transfer beyond a
single generator.

\section{Discussion and Conclusion}
\label{sec:discussion}

FinReportBench connects measurement, diagnosis, and improvement for
institution-grade financial report generation. Its results show that
professional quality is a staged decision: an artifact must first be
deliverable, then recognizable as the target document genre, and
finally complete in its institutional components. This hierarchy
explains why fluent outputs can remain unusable in practice, while the
itemized rubric turns recurrent failures into concrete targets for
skill distillation. The locked evaluation further shows that these
diagnostics can be converted into reusable constraints that improve
report identity and institutional completeness across model families.

\bibliography{references}

\end{document}